%% file: main.tex
\documentclass[letterpaper, 10 pt, conference]{ieeeconf}
\IEEEoverridecommandlockouts
\usepackage{cite}
\usepackage{amsmath,amssymb,amsfonts}
\usepackage{algorithmic}
\usepackage{graphicx}
\usepackage{textcomp}
\usepackage{xcolor}
\usepackage{graphics}
\usepackage{epsfig}
\usepackage{mathtools}
\usepackage{orcidlink}
\usepackage{caption}
\usepackage{subcaption}
\usepackage{bm}
\usepackage{url}
\usepackage{booktabs}
\def\BibTeX{{\rm B\kern-.05em{\sc i\kern-.025em b}\kern-.08em
    T\kern-.1667em\lower.7ex\hbox{E}\kern-.125emX}}

\title{Assessing Similarity Measures for the Evaluation of Human-Robot Motion Correspondence\\
}

\author{Charles Dietzel$^{1}$\orcidlink{https://orcid.org/0000-0003-1597-7853} and Patrick J. Martin$^{2}$\orcidlink{https://orcid.org/0000-0002-5896-828X}
\thanks{This research was funded in part by the Commonwealth Cyber Initiative, https://cyberinitiative.org.}
\thanks{$^{1}$Charles Dietzel is with the Commonwealth Center for Advanced Manufacturing, Disputanta, VA 23842, USA
        {\tt\small charles.dietzel@ccam-va.com}}%
\thanks{$^{2}$ Patrick J. Martin is with the Department of Computer Science, University of Richmond, VA 23173, USA
        {\tt\small patrick.martin@richmond.edu}}%
}

\begin{document}

\maketitle
\thispagestyle{empty}
\pagestyle{empty}

\begin{abstract}
One key area of research in Human-Robot Interaction is solving the human-robot correspondence problem, which asks how a robot can learn to reproduce a human motion demonstration when the human and robot have different dynamics and kinematic structures.
Evaluating these correspondence problem solutions often requires the use of qualitative surveys that can be time consuming to design and administer.
Additionally, qualitative survey results vary depending on the population of survey participants.
In this paper, we propose the use of heterogeneous time-series similarity measures as a quantitative evaluation metric for evaluating motion correspondence to complement these qualitative surveys.
To assess the suitability of these measures, we develop a behavioral cloning-based motion correspondence model, and evaluate it with a qualitative survey as well as quantitative measures.
By comparing the resulting similarity scores with the human survey results, we identify Gromov Dynamic Time Warping as a promising quantitative measure for evaluating motion correspondence.
\end{abstract}


\input{introduction}

\input{relatedwork}

\input{similaritymeasures}

\input{train_and_test}

\input{results}

\input{conclusion}

\section*{Acknowledgment}
The authors would like to thank our dance and choreography collaborators, Amelia Virtue and Kate Sicchio, for supporting the design and collection of the human demonstration data.

\bibliographystyle{IEEEtran}
\bibliography{ICRA2025_Similarity.bib}

\end{document}

%% file: introduction.tex
\section{Introduction}\label{sec:intro}

Improved sensing, perception, and control algorithms enable mobile robots to get closer to humans in a variety of domains, such as healthcare \cite{esterwood_systematic_2021}, search and rescue \cite{akgun_using_2020,wagner_robot-guided_2021}, advanced manufacturing \cite{papanastasiou_towards_2019,chen_real-time_2022}, as well as arts and entertainment \cite{herath_evokingagency_2012,jochum_tonight_2019,martin_towards_2022}.
In these scenarios, it will be necessary to incorporate non-verbal communication to improve the quality of human-robot interaction \cite{saunderson_how_2019}.
For example, non-verbal communication would support human-robot cooperative manufacturing tasks in loud manufacturing facilities where voices cannot be easily heard \cite{bergman_human-cobot_2019}, or communicating robot intent to enable new categories of artistic human-robot performances. 

Imitation learning is one approach to teach robots these non-verbal capabilities using human demonstrations \cite{hussein_imitation_2017}.
In cases where humans are performing demonstrations by moving parts of their body, e.g. motion capture, computer vision, accelerometers, it is necessary to map these demonstrated trajectories onto a robot's configuration.
Since robots will rarely have a one-to-one correspondence with the human demonstration, they require motion mapping methods that account for these embodiment differences.
This challenge is known as the \textit{correspondence problem}, e.g. \cite{alissandrakis_action_2006, liu_skill_2020}.
In recent years, multiple approaches have been proposed for solving the human-robot motion correspondence problem, such as inverse kinematics \cite{gielniak_spatiotemporal_2011, indrajit_development_2013, koenemann_real-time_2014}, re-targeting \cite{choi_towards_2019,otani_adaptive_2017}, or whole-body control \cite{arduengo_human_2021}.
Many of these approaches assume the robot possesses humanoid features and that only the robot's end effector positions matter for motion correspondence.

Creating more expressive motions would need to consider additional parts of the robot, besides the end effector position, particularly if the robot is not a humanoid.
Behavioral cloning (BC) methods, first introduced in \cite{michie_cognitive_1990}, may provide additional capabilities to learn a mapping between expert demonstrations and robot actions.
One advantage of BC models is that they may learn a mapping from the entire human demonstration vector to the robot joint vector. 
This property allows human motion to be mapped onto a robot with differing kinematic structure. 

Most correspondence solutions use qualitative surveys and measures, such as the Godspeed Questionnaire \cite{bartneck_measurement_2009}, human-robot trust \cite{charalambous_development_2016, yagoda_you_2012}, or acceptance \cite{kim_robot_2012}, to evaluate the effectiveness of the correspondence. 
These assessments provide important subjective insights into human sentiment during human-robot interaction.
However, these surveys are sensitive to the sampled population as well as experimental specifics.
It would be valuable to have quantitative measures that complement the subjective results, by objectively measuring similarity.
Recently developed heterogeneous time-similarity measures, e.g. \cite{cohen_aligning_2021, vayer_time_2022}, show promise to provide these quantitative comparisons of motions that are generated by different kinematic structures.

\begin{figure}[h]
    \centering
    \begin{subfigure}[b]{0.4\textwidth}
        \centering
        \includegraphics[width=\textwidth]{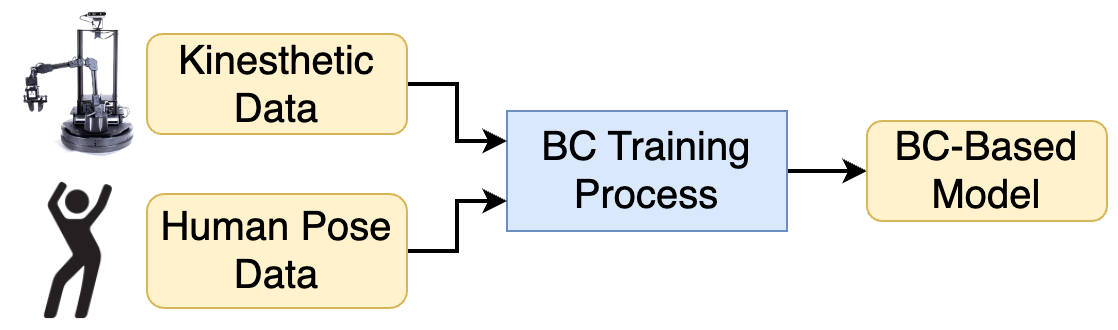}
        \caption{}
        \label{fig:training}
    \end{subfigure}
    \begin{subfigure}[b]{0.41\textwidth}
        \centering
        \includegraphics[width=\textwidth]{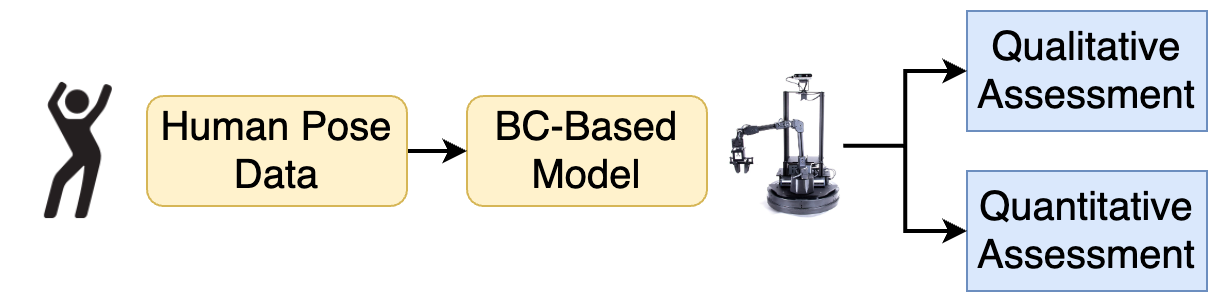}
        \caption{}
        \label{fig:deployeval}
    \end{subfigure}
    \caption{(a) The first stage of our approach collects kinesthetic and human-pose data to train behavioral-cloning based models. (b) The second stage deploys the resulting model from (a) onto the physical robot and evaluates the output qualitatively and quantitatively. }
    \label{fig:tech-approach}
\end{figure}

In this paper, we study the suitability of these measures to compute the similarity between human and robot motion by comparing the agreement between subjective survey results and the computed metrics.
As part of this effort, we also explore the applicability of BC-based models to to learn human-robot motion correspondences from kinesthetic \textit{and} human pose data. 
Our technical approach is visualized in Fig. \ref{fig:tech-approach}.

The first phase, Fig. \ref{fig:training}, applies behavioral cloning (BC) to learn a motion correspondence by using a dancer's kinesthetically taught robot motion and the dancer's body pose data example to learn a mapping onto the robot's configuration.
The resulting BC model is deployed onto the robot, Figure \ref{fig:deployeval}, and new human pose data is supplied into this model
The model produces new robot trajectories that execute on the robot platform.
We use new time-series similarity measures to quantitatively evaluate the similarity of the human pose data and generated robot trajectories.
Furthermore, we capture videos of the robot's motions and use those videos as part of a qualitative analysis by human participants.
We examine both quantitative and qualitative results to see which similarity measures best align with user qualitative data.
In summary, our contributions for this paper are:
\begin{enumerate}
    \item An application of heterogeneous time-series similarity metrics to assess human-robot motion correspondence solutions.
    \item The development of a new human-robot correspondence solution using implicit BC techniques and a novel combination of kinesthetic and human pose demonstrations.
    \item A holistic examination of these quantitative measures in the context of qualitative surveys to determine which measure might be suitable for motion correspondence evaluation.
\end{enumerate}

The organization of this paper is as follows.
In Section \ref{sec:relatedwork}, we present related work on human-robot motion correspondence mapping and time series similarity measures.
Section \ref{sec:time-series-similarity} discusses requirements for assessing human-robot motion correspondence as well as the selected the time-series similarity measures used in this study.
Our data collection and model evaluation are discussed in Section \ref{sec:train_and_test} and our results and analysis are presented in Section \ref{sec:results}.
We provide some final analysis and ideas for future research in Section \ref{sec:conclusion}.

%% file: relatedwork.tex
\section{Related Work}\label{sec:relatedwork}

This section summarizes some recent correspondence problem solutions and provides background on time-series similarity measures.

\subsection{Correspondence Problem Solutions}\label{subsec:imitation-learning}

A variety of approaches exist for solving the correspondence problem.
Inverse kinematics (IK) methods generate motion correspondences by pairing a structural model of the human body with a humanoid robot.
The spatiotemporal correspondence (STC) technique developed in \cite{gielniak_spatiotemporal_2011} directly maps human-to-robot keypoint positions and uses an optimization-based post-processing step.
The method proposed in \cite{indrajit_development_2013} performs inverse kinematics directly on human pose estimation data to generate human joint angles that are used as robot actuator commands.
Another recent IK method \cite{koenemann_real-time_2014} maps human keypoint positions to proportionally equivalent robot keypoint positions and uses inverse kinematics to compute the final robot joint commands.

Retargeting is another approach to generate motion on robots from human demonstrations, such as motion capture data.
The technique in \cite{choi_towards_2019} parameterizes and optimizes the motion transfer from human to a humanoid robot model such that the motion constraints are satisfied based on a manual selection of joints on the robot.
The approach proposed in \cite{otani_adaptive_2017} uses quadratic programming to retarget human demonstration data, but operating within a constrained environment due to robot contacts. 

The recent work in \cite{arduengo_human_2021} provides more flexibility than the prior approaches.
The authors developed a whole-body control framework that maps $n$-dimensional human poses to some $m$-dimensional robot configuration, with $m\leq n$.
This approach allows motion to be mapped from a human to a non-humanoid mobile manipulator robot, but it does require the robot manipulator to have a similar kinematic structure to a human arm. 

Alternatively, the approach presented in this paper applies the recent work in Implicit Behavioral Cloning (IBC) \cite{florence_implicit_2021,bianchini_generalization_2022}.
IBC uses energy-based models (EBM) and has demonstrated state-of-the-art performance in several offline reinforcement learning benchmarks.
Such EBMs exhibit improved data efficiency and generalization capability that would make them good candidates for solving the correspondence problem for robots that do not have a one-to-one correspondence to human demonstration data.

\subsection{Time-Series Similarity Measures}\label{subsec:time-series_similarity}

Dynamic Time Warping (DTW) \cite{sakoe_dynamic_1978} is a classical approach that maps the indices from each sequence to one or more indices from the other sequence by minimizing a distance metric at each step. 
The similarity between two time-series data sequences may be calculated even if the sequences have time shifts or scaling.

There have been multiple variations of DTW developed, such as Soft-DTW \cite{blondel_differentiable_2021}, which replaces DTW's minimum cost function with a soft-minimum operator allowing Soft-DTW to produce smoother and more accurate alignments.
Another alternative to DTW is presented in \cite{cuturi_fast_2011}, called Global Alignment Kernels (GAK).
GAK makes use of a kernel based approach to handle the time-series alignment task, increasing computational efficiency as well as improving the quality of the generated time-alignment.

Another classical similarity measure approach is Longest Common Subsequence (LCSS) \cite{vlachos_discovering_2002}.
LCSS works by attempting to identify the most similar regions between two signals and performing time-alignment on those regions.
A similarity score is generated by computing the ratio between the length of the longest common subsequence and the length of the shortest original signal. 

One challenge when comparing time series data is dealing with data that has different dimensions.
Canonical Time Warping (CTW) \cite{zhou_canonical_2009} handles this challenge by applying DTW \cite{sakoe_dynamic_1978} and Canonical Correlation Analysis \cite{kim_canonical_2009} until convergence is achieved.
The approaches in \cite{duchene_similarity_2004, tapinos_method_2013} develop variants of the longest common subsequence (LCSS) approach from \cite{vlachos_discovering_2002} to also handle this dimensionality challenge.

%% file: similaritymeasures.tex
\section{Similarity Measures for Motion Correspondence}\label{sec:time-series-similarity}

The goal of the correspondence problem is to find a mapping from a higher dimensional human motion to a corresponding lower dimensional robot motion that is as \emph{similar} as possible.
We propose the use of newly developed heterogeneous time-series similarity measures to evaluate solutions to this correspondence problem.

\subsection{Desired Similarity Measure Properties}

Time-series similarity measures have two useful properties that support the evaluation of correspondence problem solutions.
First, time-series similarity measures avoid the subjectivity issues, statistical error, financial cost, and time cost introduced by the use of human surveys.
Second, such time-series similarity measures are general enough to apply to any correspondence problem solution, regardless of motion types being generated or the robot's kinematic structure.

One limitation of the similarity measures described in Section \ref{subsec:time-series_similarity} is that they compare pairs of time series data with \emph{identical dimensions}. 
But our goal is to identify and apply similarity measures that compute the degree of similarity between a human motion and any corresponding robot motion.
Given that robots will have different kinematic structures, the data structure of each robot motion will differ from its corresponding human motion data.
This mismatch requires similarity measures that handle \textit{heterogeneous} data sets with different dimensions.

Another limitation of the measures in Section \ref{subsec:time-series_similarity} is that they struggle with non-linearly and spatially distorted data sets.
This issue is problematic for evaluating correspondence problem solutions since the mapping between human and robot time series trajectories is generally non-linear.
Therefore, we examine two recent time-series similarity measures capable of handling heterogeneous, non-linearly distorted data: Gromov DTW (GDTW) \cite{cohen_aligning_2021} and Global Invariances (DTW-GI) \cite{vayer_time_2022}.

\subsection{Candidate Similarity Measures}

GDTW improves upon the flexibility and accuracy of the original DTW formulation is several ways.
The authors connect that alignment matrices in the DTW objective can be treated as analogues of the coupling matrices that appear in Wasserstein distance from the classical optimal transport problem \cite{villani_optimal_2009}.
The authors augment their algorithm formulation by replacing Wasserstein distance with Gromov–Wasserstein distance \cite{memoli_gromovwasserstein_2011}, which allows GDTW to be invariant to various isometries, including rotation and translation, allowing it to compare heterogeneous time-series data.
The authors of GDTW also developed a variant of GDTW that uses SoftDTW to smooth and soften the generated time-series alignment, resulting in a greater robustness to signal noise.

Alternatively, DTW-GI performs joint optimization over temporal alignments and feature space transformations.
The temporal alignment step is performed using DTW, while the feature space transformation can be any combination of rigid transforms or even other families of transforms.
This customizable feature space transformation allows DTW-GI to remain invariant to rotation, translation, and other types of transforms, which provides flexibility when evaluating time-series pairs with complex transformations.
The authors also also developed a variant that uses SoftDTW.

Given the advantageous properties of these two time-series similarity measures, we apply GDTW and DTW-GI (as well as their SoftDTW variants) to evalue human-robot motion correspondence solutions. 
They posses the greatest capacity to evaluate similarity among heterogeneous time-series data that has undergone significant non-linear transformation.

%% file: train_and_test.tex
\section{Model Training and Evaluation}\label{sec:train_and_test}

To determine the usefulness of the selected time-series similarity measures as a means of evaluating correspondence problem solutions, we implemented an Implicit Behavioral Cloning (IBC) \cite{florence_implicit_2021} approach as the basis for our correspondence problem solution.
As discussed in Section \ref{sec:relatedwork}, IBC makes use of energy based models (EBMs) that have good data efficiency as well as better to out-of-distribution data extrapolation \cite{florence_implicit_2021,bianchini_generalization_2022}.
EBMs also perform better than feedforward models at trajectory modeling tasks \cite{du_implicit_2020} and achieve
near SOTA results on a number of well known benchmarks including the D4RL Human-Experts (Franka and Adroit) tasks.
We selected a traditional feedforward Multilayer Perceptron (MLP) network as a baseline for comparison with the IBC approach.

To train and evaluate the IBC model and baseline, we developed a dataset of human motion trajectories and corresponding robot joint trajectories.
These trajectories were designed and performed by our dancer and choreographer collaborators, each of whom possessed an expert understanding of human body dynamics as well as the skill set to imagine how human motions might be represented in the kinematic structure of a non-humanoid robot.
After training each of these models using our collected dataset, we evaluated them using qualitative surveys taken by dancers and choreographers as well as engineers.
We also evaluated each model's outputs using the GDTW and DTW-GI time-series similarity measures.

\subsection{Data Collection}\label{sec:data_collection}

\begin{figure}[ht]
\centering
\includegraphics[width=0.7\columnwidth]{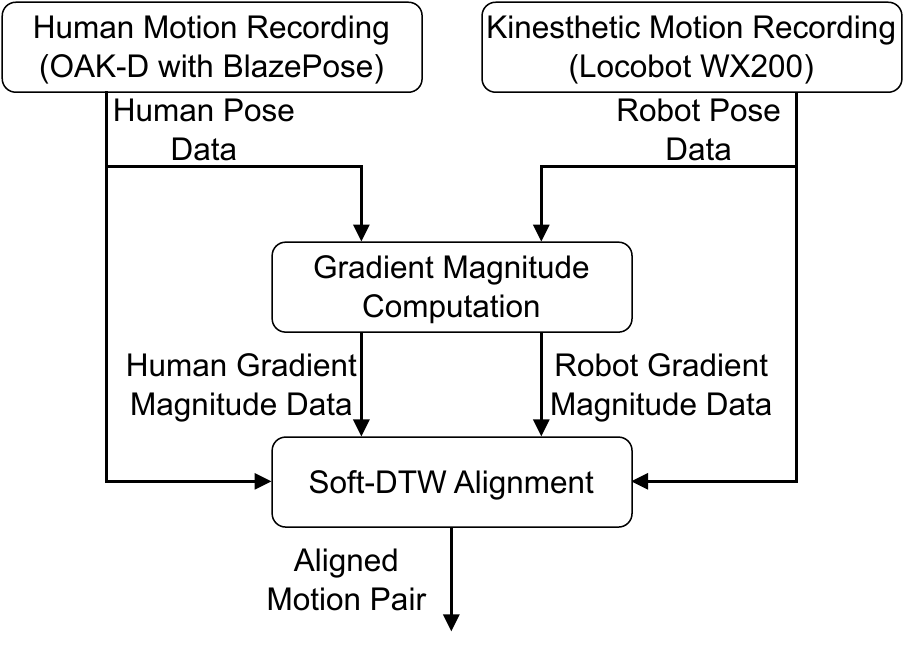}
\caption{The data collection process that was used to build our data set for training and testing.}
\label{fig:data_collection}
\end{figure}

Each motion was designed to be as dissimilar as possible from the others to provide a wide variety of trajectories for model training.
The full process for capturing and pre-processing the human and robot motion data is illustrated in Figure \ref{fig:data_collection}.

\begin{figure}[h]
    \centering
    \begin{subfigure}[b]{0.12\textwidth}
        \centering
        \includegraphics[width=\textwidth]{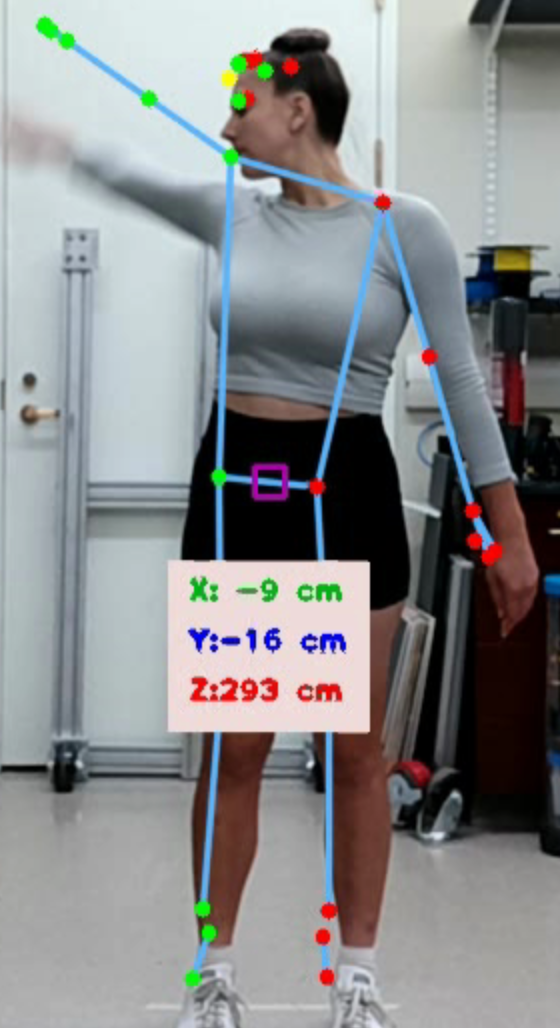}
        \caption{}
        \label{fig:amelia1}
    \end{subfigure}
    \begin{subfigure}[b]{0.15\textwidth}
        \centering
        \includegraphics[width=\textwidth]{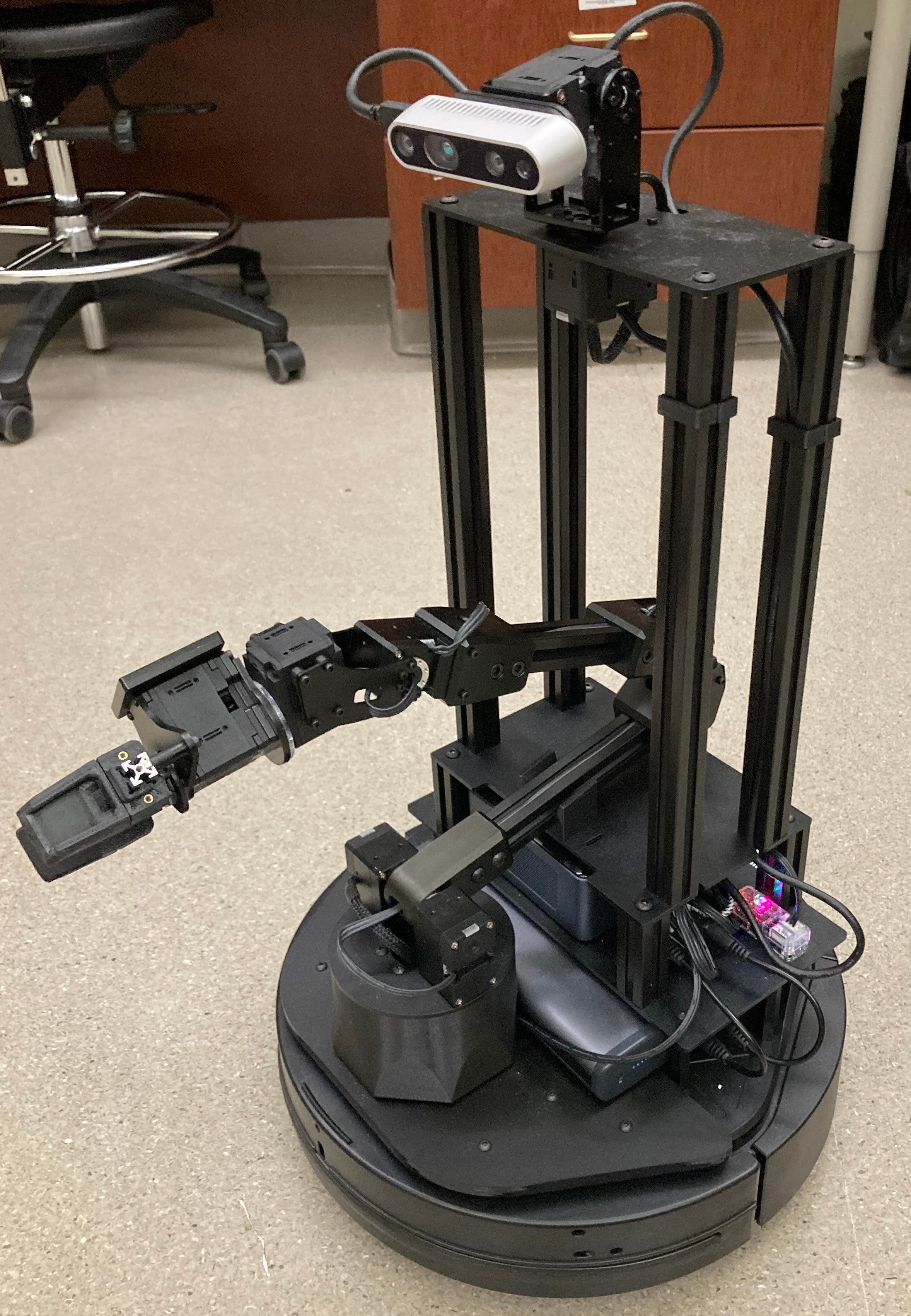}
        \caption{}
        \label{fig:isadora}
    \end{subfigure}
    \caption{(a) An example output image from our human-pose data source and (b) the target non-humanoid robot.}
    \label{fig:pecm_platforms}
\end{figure}

The process collects two sources of information: 1) the keypoints of a human motion, captured by an OAK-D stereo depth camera loaded with the BlazePose human pose estimation model \cite{bazarevsky_blazepose_2020}, shown in Figure \ref{fig:amelia1}, and 2) the joint angles of a 5 degree-of-freedom arm attached to a Trossen Robotics LoCoBot WX200 mobile manipulator robot, shown in Figure \ref{fig:isadora}.
The dancer's motion demonstratration was collected with the OAK-D camera.
Subsequently, the dancer kinesthetically demonstrated a corresponding motion using the LoCoBot's manipulator, which captured the data using its built in servo motor angle sensors.
Throughout this robot trajectory recording process, the dancer was instructed to move the manipulator arm in a way that they felt \emph{represented} the same motion they performed.
Our training dataset consisted of 11 different corresponding human-robot motion pairs, with 6 demonstration samples of each motion. 
Five additional motion pairs were collected for the test set, with one demonstration each.
Since each motion was about 10-15 seconds long and the recordings were captured at 10Hz, this added up to roughly 10000 human-robot joint position pairs between the training and test data.

After the motion data was collected, it was augmented by generating mirrored versions of each human and robot motion pair by flipping the left and right sides of the human body and horizontally reflecting the robot's joint positions.
All dancer keypoint data was reduced to their shoulder, elbow, wrist, palm, and finger keypoints.
Finally, time-alignment was applied to each human and robot motion data pair.
Given the construction of the data collection scenario, we assume there is an \textit{intended} correspondence between each pair of human-robot trajectories. 
Based on this assumption we compute the gradient magnitude of each corresponding pair of human and robot trajectories and use this information, as well as the original demonstration data, to time-align the entire dataset using Soft-DTW \cite{blondel_differentiable_2021}.

\subsection{Training}\label{sec:training}

We trained our IBC correspondence model using the EBM-based training process.
We trained our baseline neural network with a mean squared error (MSE) loss function and standard backpropagation.
Both of these models use a multilayer perceptron (MLP) architecture.
In the training process, we performed hyperparameter grid search using the GDTW, Soft GDTW, DTW-GI, and Soft DTW-GI similarity measures described in \cite{cohen_aligning_2021, vayer_time_2022} as loss functions.
The model with the highest average similarity between each human test motion and the corresponding robot joint trajectory generated by that model was selected for further evaluation.
Hyperparameters for each of the four selected models is shown in Table \ref{tab:PECM_params} and Table \ref{tab:MLP_params}.

\begin{table}[ht]
    \caption{IBC Model Hyperparameters}
    \label{tab:PECM_params}
    \resizebox{\columnwidth}{!}{\begin{tabular}{lrrr}
        \toprule
        & IBC-1 & IBC-2 & Searched Values\\
        Hyperparameter & & &\\
        \midrule
        Training iterations & 100000 & 100000 &\\
        Batch size & 512 & 512 &\\
        Sequence length & 3 & 3 &\\
        Network size (width x depth) & 512x4 & 512x4 & 512x4, 256x8, 128x16\\
        Activation function & ReLU & ReLU &\\
        Learning rate & 1e-3 & 1e-3 & 1e-3, 5e-4, 1e-4\\
        Learning rate decay & 0.99 & 0.99 &\\
        Learning rate decay steps & 100 & 100 &\\
        Gradient penalty & Final step only & Final step only &\\
        Gradient margin & 1 & 1 &\\
        Training counter-examples & 8 & 4 & 8, 4\\
        Langevin iterations & 100 & 100 &\\
        Langevin learning rate init. & 0.1 & 0.1 &\\
        Langevin learning rate final & 1e-5 & 1e-5 &\\
        Langevin polynomial decay power & 2 & 2 &\\
        Langevin delta action clip & 0.1 & 0.1 &\\
        \midrule
        Selected by & Soft DTW-GI & GDTW, Soft GDTW, DTW-GI & \\
        \bottomrule
    \end{tabular}}
\end{table}

\begin{table}[ht]
    \caption{MLP Baseline Hyperparameters}
    \label{tab:MLP_params}
    \resizebox{\columnwidth}{!}{\begin{tabular}{lrrr}
        \toprule
        & MLP-1 & MLP-2 & Searched Values\\
        Hyperparameter & & &\\
        \midrule
        Training iterations & 100000 & 100000 &\\
        Batch size & 512 & 512 &\\
        Sequence length & 3 & 3 &\\
        Network size (width x depth) & 512x4 & 512x4 & 512x4, 256x8, 128x16\\
        Activation function & ReLU & ReLU &\\
        Learning rate & 1e-3 & 5e-4 & 1e-3, 5e-4, 1e-4\\
        Learning rate decay & 0.99 & 0.99 &\\
        Learning rate decay steps & 100 & 100 &\\
        Dropout rate & 0.1 & 0.1 & 0, 0.1\\
        \midrule
        Selected by & Soft DTW-GI & GDTW, Soft GDTW, DTW-GI & \\
        \bottomrule
    \end{tabular}}
\end{table}

Among all four similarity measures, three of them selected the same set of hyperparameters for each model, resulting in two hyperparameter sets for both models. 
Additionally, the two sets of hyperparameters varied only by the number of IBC training counter-examples and by the MLP baseline's learning rate. 
After these similarity measures identified hyperparameter combinations for each model, we selected the best models and evaluated their human-robot motion correspondence.

\subsection{Quantitative Metrics}

Quantitative evaluation of these models used the GDTW, Soft GDTW \cite{cohen_aligning_2021}, DTW-GI, and Soft DTW-GI \cite{vayer_time_2022}, with CTW \cite{zhou_canonical_2009} used as a baseline for comparison. 
We selected CTW as a baseline because it is a well known measure that is capable of evaluating similarity of heterogeneous time-series data, but is still sensitive to rotation and translation. 
The IBC and baseline models in Tables \ref{tab:PECM_params} and \ref{tab:MLP_params} were deployed on the LoCoBot to generate \textit{new} robot motion trajectories using the five unseen human test motions collected in Section \ref{sec:data_collection}.
The similarity between each human test motion and their generated robot motion was calculated using each similarity measure. 
We averaged the similarity for each model and each similarity measure across all five motions, as shown in Figure \ref{fig:quantitative_eval}.

\subsection{Qualitative Surveys}\label{sec:qual_surveys}

We complemented the quantitative measures with qualitative surveys to assess if any (or, all) of the quantitative similarity metrics align with the humans' subjective notion of similarity.
The selected models from Section \ref{sec:training} generated robot test motion trajectories corresponding to three of the five human test motions collected in Section \ref{sec:data_collection}.
We included only three randomly selected test motions to ensure that the survey was of a manageable length for our group of participants, which consisted of five dancers/choreographers and seven engineers.
With three test motions and with motions generated by four different neural network models plus the one human demonstrated motion, all survey participants were asked to rate the same 15 motion pairs. 
Additionally, a video was recorded of each generated robot motion.
In the survey, participants were shown video of each human motion followed by a corresponding robot motion and asked to rate these robot motions on a variety of criteria.
The survey randomized the presentation order of the motion pairs under study.

\begin{table}[ht]
    \caption{Survey Questions}
    \label{tab:survey_questions}
    \resizebox{\columnwidth}{!}{\begin{tabular}{|p{0.2cm}|p{5cm}||p{1.7cm}|p{1.7cm}|}
        \hline
        \textbf{\#} & \textbf{Question} & \textbf{1 Rating} & \textbf{5 Rating}\\
        \hline
        1. & How well do you think the robot motion replicated the movement of the human motion? & Not Well & Very Well\\
        \hline
        2. & How well do you think the robot motion replicated the style of the human motion? & Not Well & Very Well\\
        \hline
        3. & Please rate your impression of the robot on this scale: & Fake & Natural\\
        \hline
        4. & Please rate your impression of the robot on this scale: & Machinelike & Humanlike\\
        \hline
        5. & Please rate your impression of the robot on this scale: & Unconscious & Conscious\\
        \hline
        6. & Please rate your impression of the robot on this scale: & Artificial & Lifelike\\
        \hline
        7. & Please rate your impression of the robot on this scale: & Moving rigidly & Moving elegantly\\
        \hline
        8. & Please rate your impression of the robot on this scale: & Choppy & Smooth\\
        \hline
        9. & Please rate your impression of the robot on this scale: & Unbalanced & Even\\
        \hline
    \end{tabular}}
\end{table}

Table \ref{tab:survey_questions} shows the survey questions, each of which asked participants to rate the robot's motion performance on an integer scale from 1 (low) to 5 (high).
Questions 1 and 2 attempted to directly measure the quality of the correspondence between each human motion and robot motion.
Questions 3 through 7 were adapted from the anthropomorphism section of the Godspeed Questionnaire \cite{bartneck_measurement_2009}.
Questions 8 and 9 were designed to assess the human subjects' interpretation of smoothness of each generated motion.

%% file: results.tex
\section{Model Analysis}\label{sec:results}

Our primary goal of this work is to determine the suitability of GDTW and DTW-GI for evaluating human-robot motion correspondence quality.
A secondary goal is to evaluate the performance of the IBC model and to determine its ability to solve the correspondence problem solution compared to a baseline. 

\subsection{Qualitative Results}

Table \ref{tab:survey_results} shows the qualitative survey results described in Section \ref{sec:qual_surveys}.
The top row of the table shows the how the participants rated motions that were the result of the dancer's kinesthetic demonstration on the robot arm.
The survey results show that neither IBC model outperformed the MLP baseline models.
The IBC models performed the worst relative to the baseline in Q8 and Q9, with likert scale results that were roughly 0.75 points lower.
This indicates that the IBC's most severe deficiency was in generating motions that were smooth and balanced.
IBC also performed worse than the baseline when evaluated on Q3-Q7, which cover the anthropomorphism aspects of the Godspeed Questionnaire.

\begin{table}[ht]
\caption{Survey Results (higher is better)}
\label{tab:survey_results}
\resizebox{\columnwidth}{!}{\begin{tabular}{lrrrrrrrrrr}
\toprule
 & Q1 & Q2 & Q3 & Q4 & Q5 & Q6 & Q7 & Q8 & Q9 & Average \\
Motion Source &  &  &  &  &  &  &  &  &  &  \\
\midrule
Dancer Demonstration & 3.67 & 3.89 & 3.78 & 3.50 & 3.25 & 3.64 & 3.39 & 3.72 & 3.83 & 3.63 \\
IBC-1 & 2.53 & 2.61 & 2.36 & 2.31 & 2.42 & 2.50 & 2.44 & 2.44 & 2.50 & 2.46 \\
IBC-2 & 2.56 & 2.64 & 2.53 & 2.39 & 2.44 & 2.39 & 2.31 & 2.42 & 2.25 & 2.44 \\
MLP-1 & 2.53 & 2.69 & 3.00 & 2.61 & \textbf{2.61} & 2.75 & 2.61 & 2.97 & 3.11 & 2.77 \\
MLP-2 & \textbf{2.69} & \textbf{3.08} & \textbf{3.17} & \textbf{2.72} & 2.47 & \textbf{2.83} & \textbf{2.78} & \textbf{3.33} & \textbf{3.17} & \textbf{2.92} \\
\bottomrule
\end{tabular}}
\end{table}

Interestingly, the area in which the IBC models compared \textit{most} favorably with the baseline is Q1 and Q2, which asked survey participants about the robot motion quality and human motion correspondence.
IBC achieved near parity with the baseline, achieving equal or better results than one of the baseline models.
These results suggest that the biggest weakness of IBC generated motions was their choppy generated motion.
One potential explanation for this is that, as described in \cite{florence_implicit_2021}, IBC is more sensitive to data quality than traditional feedforward models.
Any errors present in the OAK-D human pose recordings and/or in the kinematic robot demonstrations would therefore have had a greater negative effect on IBC than on the baseline.

\subsection{Quantitative Results}

\begin{figure}[ht]
    \centering
    \includegraphics[scale=0.4]{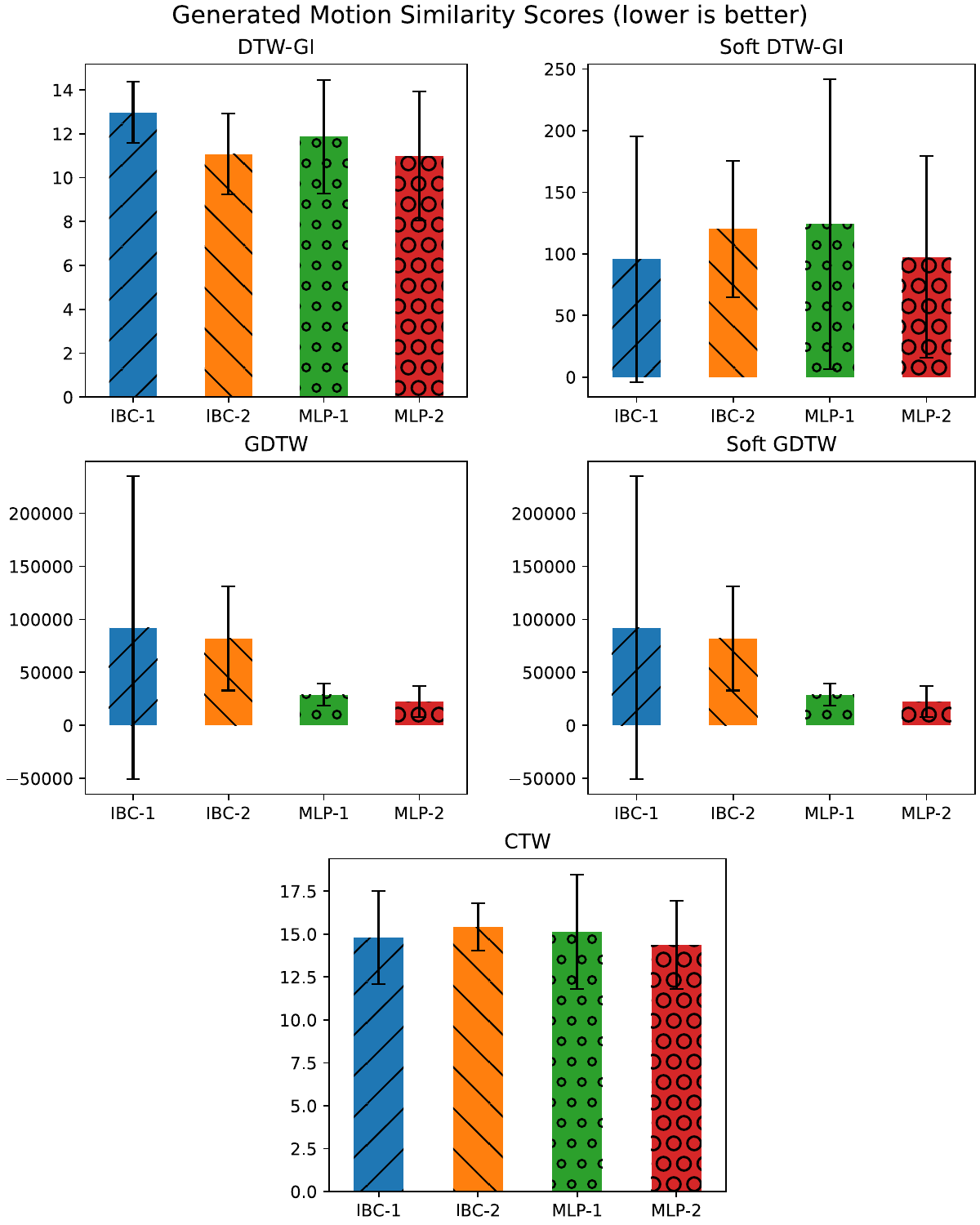}
    \caption{This figure shows the average time-series similarity scores across the IBC and MLP baseline models. The black error bars show the standard deviation of the results. 
    }
    \label{fig:quantitative_eval}
\end{figure}

Figure \ref{fig:quantitative_eval} shows the quantitative scores for each model using the four different time series similarity measures as well as the CTW baseline technique.
We note that the similarity scores for the IBC models were the same or slightly worse than the MLP baseline.

The baseline similarity measure CTW failed to meaningfully differentiate results between the different models, producing near identical scores for each.
CTW works by fitting a rigid transformation between two time-series inputs.
The transformation between human keypoint positions and robot joint angles is more complex and nonlinear than a rigid transform, so CTW proved ineffective as an evaluation method for this task.

On the other hand, GDTW and Soft-GDTW both produced good results by illustrating an observable disparity between the IBC models and the MLP baseline models.
These measures also produced close results when comparing the model types themselves, e.g. IBC-1 compared to IBC-2.
This ability to successfully differentiate between the IBC and MLP models implies that GDTW and Soft-GDTW were both able to identify differences in the quality of motion correspondence generated by each category of model.

Neither DTW-GI nor Soft DTW-GI performed as well as GDTW, with a much smaller disparity between similarity scores for the IBC and MLP models. 
Also, the scores for both IBC models were relatively different, and so were the scores for the two MLP models. 
These weaknesses reduce the potential usefulness of the DTW-GI approaches for evaluating the quality of motion correspondence.

\subsection{Qualitative and Quantitative Results Comparison} 

GDTW and Soft-GDTW show promise in evaluating human-robot motion correspondence. 
Additionally, by comparing GDTW's similarity scores with the human survey results for each model, we see that 
the results of GDTW assign better scores to the baseline models and worse scores to the IBC models.
This agreement between the GDTW similarity measures and the qualitative human survey responses provides evidence that these similarity measures provide quantitative data that complements subjective evaluations. 
This agreement between GDTW and the survey also implies that GDTW numerically captured meaningful aspects of the human-robot motion correspondence quality.

%% file: conclusion.tex
\section{Conclusion}\label{sec:conclusion}

In this paper, we proposed and studied the use of recently developed time-series similarity measures to quantitatively evaluate the degree of correspondence between higher dimensional human keypoint trajectories and lower dimensional, non-anthropomorphic robot joint trajectories.
Additionally, we explored the use of Implicit Behavioral Cloning (IBC) to solve the human-robot motion correspondence problem.
This model was trained using a novel combination of whole-body human pose demonstrations and robot kinesthetic demonstrations.

We evaluated this IBC model using time-series similarity measures as well as a qualitative human survey that assesses the generated motion's quality.
Our similarity scores and survey results show that IBC did not outperform our baseline feedforward MLP model, possibly due to IBC's greater sensitivity to data error.
Two of the four time-series similarity measures we considered, GDTW and Soft GDTW, agreed with our qualitative human survey results.
This result indicates that these two similarity measures would be good candidates for correspondence evaluations between human-robot motion pairs, even for non-humanoid robots.
We suggest that other human-robot motion correspondence researchers could use these similarity measures to supplement human surveys.

In our current study, we only tested these similarity measures on one robot type.
We plan to expand our work to include evaluations on robots with varying kinematic structures and with different sets of human joint keypoints in the recorded demonstrations, e.g. leg, torso, etc. 
Since these similarity measures are also differentiable, another promising area of future research would be to evaluate these similarity measures as loss functions for use in model training to generate robot motions that correspond to human motions. 
We also plan to investigate other neural network model types that might better encode the mapping over heterogeneous time series data.